%% file: emnlp2023.tex
\pdfoutput=1

\documentclass[11pt]{article}

\usepackage[]{EMNLP2023}
\usepackage{multirow}
\usepackage{times}
\usepackage{latexsym}
\usepackage{ulem}
\usepackage[T1]{fontenc}

\usepackage[utf8]{inputenc}
\usepackage{url}
\usepackage{microtype}

\usepackage{inconsolata}
\usepackage{graphicx}
\usepackage{framed}
\usepackage{array}
\usepackage{subcaption}
\usepackage{CJK}
\usepackage{makecell}
\usepackage[most]{tcolorbox}
\usepackage{cleveref} 
\newcommand{\Blue}[1]{\colorbox[RGB]{153,204,255}{#1}}
\newcommand{\Yellow}[1]{\colorbox{yellow}{#1}}
\newcommand{\Green}[1]{\textcolor[RGB]{76,153,50}{#1}}
\newcommand{\Pink}[1]{\textcolor[RGB]{210,18,150}{#1}}

\newcommand{\figref}[1]{Figure \ref{#1}}

\newcommand{\tabref}[1]{Table \ref{#1}}

\definecolor{shadecolor}{RGB}{224, 224, 224}
\definecolor{framecolor}{RGB}{160,160,160}
\newenvironment{prompt}
{\begin{center}\begin{tcolorbox}[colback=shadecolor,
                  colframe=framecolor,
                  width=7.4cm,
                  arc=2mm, auto outer arc,
                  boxrule=0.3pt,
                 ] \small}
{\end{tcolorbox}\end{center}\par}

\title{LLM-empowered Chatbots for Psychiatrist and Patient Simulation: Application and Evaluation}

\author{Siyuan Chen$^1$, Mengyue Wu$^2$\thanks{~~Corresponding Authors.}, Kenny Q. Zhu$^{3*}$,  \\ 
{\bf Kunyao Lan$^4$}, {\bf Zhiling Zhang$^5$}, {\bf Lyuchun Cui$^6$} \\
$^{1,2,3,4,5}$Shanghai Jiao Tong University, Shanghai, China \\
$^{6}$Shanghai Mental Health Center, Shanghai, China \\
\texttt{\{$^1$chensiyuan925, $^2$mengyuewu, $^4$lankunyao, $^5$blmoistawinde\}@sjtu.edu.cn,}\\
\texttt{$^3$kzhu@cs.sjtu.edu.cn, $^6$cuilvchun@outlook.com}
}

\begin{document}
\maketitle
\begin{abstract}
Empowering chatbots in the field of mental health is receiving increasing amount of attention, while there still lacks exploration in developing and evaluating chatbots in psychiatric outpatient scenarios.
In this work, we focus on exploring the potential of ChatGPT in powering chatbots for psychiatrist and patient simulation. We collaborate with psychiatrists to identify objectives and iteratively develop the dialogue system to closely align with real-world scenarios. In the evaluation experiments, we recruit real psychiatrists and patients to engage in diagnostic conversations with the chatbots, collecting their ratings for assessment. Our findings demonstrate the feasibility of using ChatGPT-powered chatbots in psychiatric scenarios and explore the impact of prompt designs on chatbot behavior and user experience.
\end{abstract}

\input{intro}

\input{prob_def}

\input{method}

\input{eval_frame}

\input{data}
\input{experiment}
\input{conclusion}

\input{ethics}

\bibliography{anthology,custom}
\bibliographystyle{acl_natbib}

\appendix

\input{appendix}
\input{related}

\end{document}

%% file: intro.tex
\section{Introduction}
\label{sec:intro}

Conversational Agents (i.e., chatbots) are becoming increasingly popular in the mental health domain \cite{sabour2022chatbots}. Applications designed for mental health therapy or coaching in daily life, such as Woebot\footnote{\url{https://woebothealth.com}} and Wysa\footnote{\url{https://www.wysa.com}}, are gaining widespread attention for their ability to reduce users' negative emotions~\cite{Grove2021Codevelop} and promote a healthy lifestyle~\cite{Fadhil2019AssistiveCA}. Another notable application is chatbot-based symptom checkers~\cite{Yue2023Beyond}, which emulate human-like conversations while assessing users' symptoms, resembling interactive questionnaires.

However, there is still limited exploration in developing and evaluating chatbots that can (i) conduct diagnosis conversations like a psychiatrist
or (ii) simulate patients in the psychiatric outpatient scenarios, though they have significant real-world applications. 
Doctor chatbots can be effective tools for mental disorder screening~\cite{pacheco2021Smart} in lieu of official medical diagnosis. Patient chatbots can serve as Standard Patients (SP) in medical education, making the process more efficient and cost-effective~\cite{Torous2021growing}.

Developing and evaluating such chatbots is particularly challenging due to the unique nature of mental health issues, including (i) the difficulty in obtaining data because of privacy concerns; (ii) the inherent ambiguity and subjectivity of mental disease symptoms. 
Moreover, relying solely on scales (e.g., PHQ-9) for mental disorder screening cannot provide trustworthy diagnosis, because in real outpatient scenario, patients often feel ashamed or afraid of disclosing their true conditions and difficult to describe their mental state objectively~\cite{Salaheddine2016Identify}. Thus, even experienced psychiatrists struggle to obtain the meaningful response from patients. 

Consequently, the design goals and conversational styles of these chatbots are different from the chatbots for mental health therapy and symptom checking.
We argue that chatbots for psychiatric diagnosis can not achieve satisfying performance by simply collecting symptoms like questionnaires. Instead, they should be equipped with various professional skills, such as emotional support, to complete the diagnosis task effectively.
What's more, patient chatbots should aim to resemble real patients more closely, rather than precisely and robotically reporting their symptoms without any emotional fluctuations.

\begin{figure*}[th]
	\centering
	\includegraphics[width=0.8\linewidth]{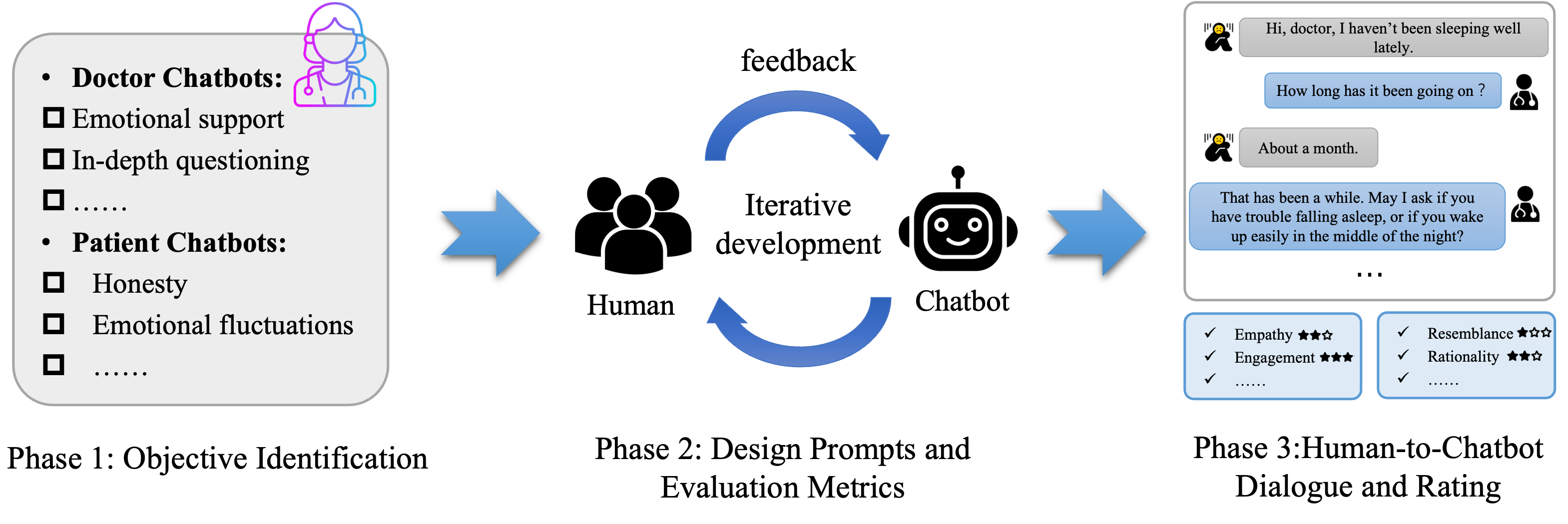}
	\caption{The overview of the psychiatrist-guided three-phase study.}
	\label{fig:pipeline}
\end{figure*}



Achieving these goals is quite difficult for conventional rule-based~\cite{Medeiros2018UsingCF,Jaiswal2019Virtual} or data-based~\cite{yao-etal-2022-d4, Fansi2022DDXPlus, Lin2021Graph} methods. Fortunately, recent advancements in large language models (LLMs), especially with the emergence of ChatGPT\footnote{\url{https://chat.openai.com/}}, provide a new way to develop chatbots that can convincingly portray specific roles.  Equipped with comprehensive training data and knowledge, LLMs can generate diverse tones and symptom descriptions with appropriate prompts rather than fine-tuning on extensive domain data.

Therefore, in this work, we aim to (i) respectively investigate the potential of ChatGPT in simulating psychiatrists and mental disordered patients in a clinical diagnosis scenario\footnote{For the sake of clarity, we will refer to these two types of chatbots as the \textbf{``doctor chatbot''} and \textbf{``patient chatbot''} respectively in the subsequent sections.}, 
as well as (ii) build a comprehensive evaluation framework for these chatbots, answering the question about what constitutes an exceptional psychiatrist chatbot and a truly patient-like chatbot.
To develop and evaluate a system that truly satisfies users' expectations, we followed a human-centered design methodology. The study consists of three phases (See Figure~\ref{fig:pipeline}).
We first collaborated with psychiatrists to identify a set of objectives for doctor and patient chatbots (\textbf{Phase 1}). 
Based on these objectives, we conducted an experimental study (\textbf{Phase 2}) to design appropriate prompts for ChatGPT-based chatbots and establish an evaluation framework that incorporates both human evaluation and automatic metrics aligned with the objectives from Phase 1. Importantly, the design of prompts and metrics was iterated based on human feedback, with each version evaluated and improved with input from psychiatrists.

Further, to better evaluate the performance of these chatbots with varying prompt designs, we recruit real psychiatrists and patients to engage in diagnostic conversations with the simulated patient and doctor chatbots, respectively, and collect their ratings after conversation (\textbf{Phase 3}). We also conduct a comparison between the behavior of real and simulated psychiatrists based on the dialogue history, which yields some interesting findings. 
The main contributions of this work are:
\begin{itemize}
    \item We formalize the task of developing doctor and patient chatbot for diagnostic purposes in a psychiatric outpatient setting.
    \item We conduct a user-centered investigation into the design and evaluation of these chatbots. Through an iterative development process, we actively sought feedback from both patients and psychiatrists, allowing us to establish a more solid and applicable chatbot system and evaluation framework. 
    \item Through detailed prompt engineering and experiments, we demonstrate the feasibility of utilizing ChatGPT-powered chatbots in professional domains that demand specialized skills or unique language style. We also use interactive human evaluation to explore how different prompt designs influence user experience.
\end{itemize}


%% file: prob_def.tex
\section{Objectives}
\label{sec:objectives}

In phase 1, we consulted with 7 psychiatrists and worked together to establish the objectives we hope the doctor and patient chatbots can achieve, which will guide us in the following stage.

Since the diagnosis standards of different mental disorders vary greatly, psychiatrists recommend concentrating on depressive disorders for this study, while leaving the scaling to include other disorders as future work.

\subsection{Doctor Chatbot}
\label{sec:doc_requirements}
As a doctor chatbot, the primary task is to conduct a professional diagnostic process for the patient and provide an accurate diagnosis. To achieve this, and to offer patients a superior healthcare experience, a good doctor chatbot should possess the following three capacities:
\begin{itemize}
    \item \textbf{Comprehensiveness:} Inquire about the key symptoms of depression, including sleep, mood, diet, and other relevant aspects that are required for diagnosis.
    \item \textbf{In-depth Questioning:} Conduct thorough questioning based on patient's responses to gain a better understanding of the symptoms.
    \item \textbf{Empathy:} Demonstrate empathy and provide emotional support towards patients' experiences to encourage them to express their situation more freely and obtain more information, which can lead to better diagnostic results.
\end{itemize}

\subsection{Patient Chatbot}
\label{sec:pat_requirements}

After establishing objectives for doctor chatbots, we encountered difficulties when defining the requirements for chatbots that resemble real patients. This is due to the fact that individuals with the same disorder can exhibit significant variations in their manifestations. Moreover, psychiatrists, though experienced, have no firsthand chatting experience with a ``non-patient-like'' chatbot, making it challenging for them to generalize the requirements for a ``patient-like'' chatbot.

To address this issue, we decide to develop an initial version of the chatbot first. This allows psychiatrists to interact with ``non-patient-like'' examples, which can help them better define the characteristics and behaviors that constitute a ``patient-like'' chatbot. Based on their feedback, we then iterate and update the chatbot accordingly.
At this phase, we only establish one fundamental requirement for a patient chatbot.
\begin{itemize}
    \item \textbf{Honesty:} 
        Provide an accurate and rational description of symptoms in its user profile, without reporting any nonexistent symptoms.
\end{itemize}

%% file: method.tex
\section{Prompt Design}
\label{sec:method}
We describe the iterative methodology of designing prompts with users' feedback, which will be listed in bullet point in this section. 


\subsection{Doctor Chatbot}

\begin{figure}[th]
	\centering
	\includegraphics[width=\linewidth]{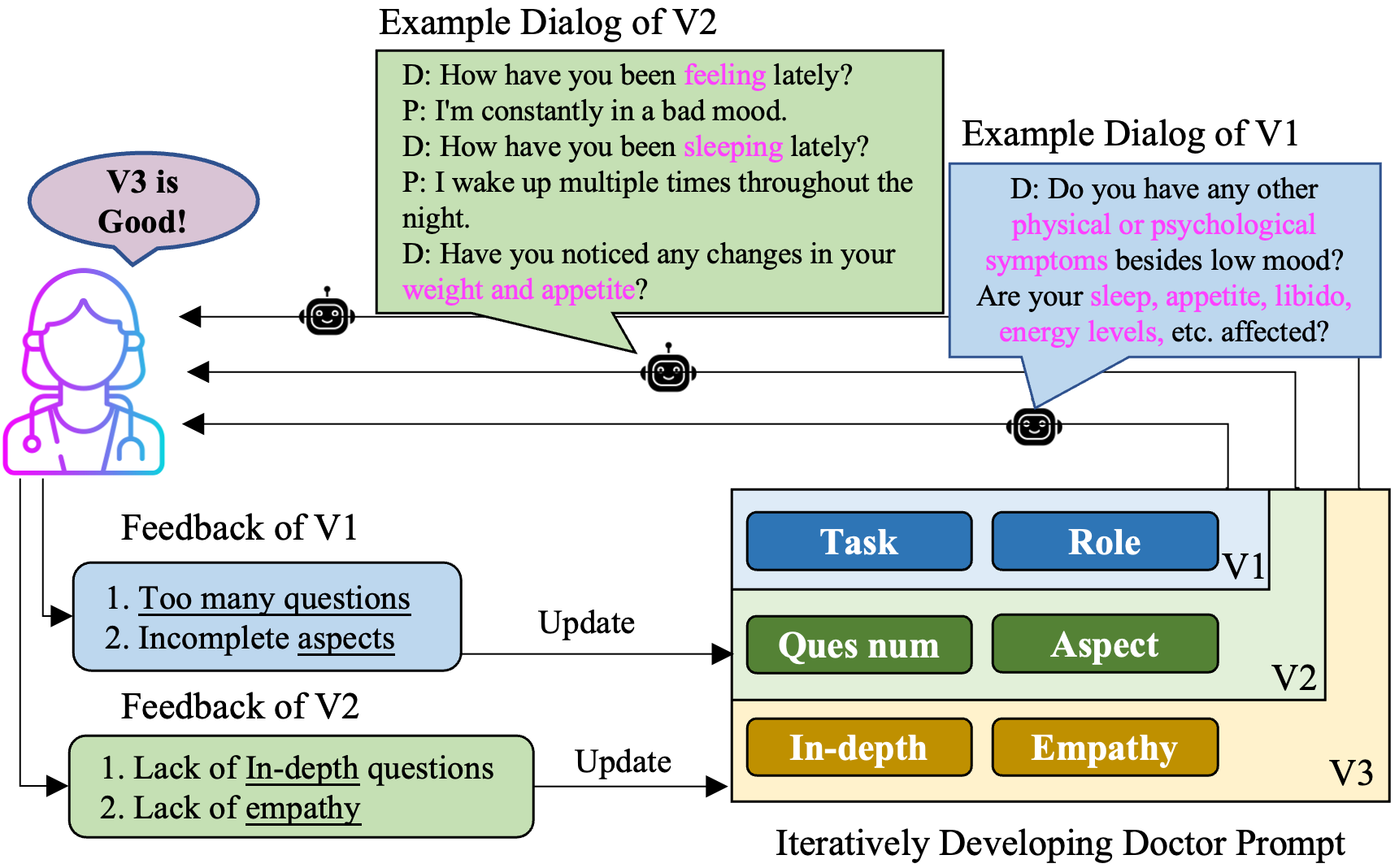}
	\caption{The iterative development process of the prompt of doctor chatbots. Psychiatrists will identify the limitations of the current version, and we will address these issues in the subsequent version.}
	\label{fig:doc_prompt_design}
\end{figure}

\label{sec:doc_prompt}
\paragraph{Version 1}
The original version of the prompt for doctor chatbot is as follows. We simply describe the task without providing any other information. 
\begin{prompt}
    \textcircled{1} Please play the \uline{role} of a psychiatrist. 
    \textcircled{2} Your \uline{task} is to conduct a professional diagnosis conversation with me based on the DSM-5 criteria. 
\end{prompt}
\begin{itemize}
    \item Although the chatbot's questions are in line with DSM-5\cite{american2013diagnostic}, it asks about almost all symptoms in one go (See Example V1 in Figure \ref{fig:doc_prompt_design}), which can be overwhelming for the patient. 
    \item Additionally, the chatbot draws a diagnosis conclusion in less than five turns without gathering sufficient information, which may result in unreliable diagnostic outcomes and fail to achieve the objective of ``comprehensiveness''.
\end{itemize}

\paragraph{Version 2}
Therefore, we developed a new version to address these problems by adding sentence \textcircled{3} and \textcircled{4} (i.e., ``ques num'' and ``aspect'' in Figure \ref{fig:doc_prompt_design}) to the original prompt. 
\begin{prompt}
    \textcircled{3} Your questions should \uline{cover at least the following aspects}: [\ldots]\protect\footnotemark. 
    \textcircled{4} Please only ask \uline{one question at a time}.
\end{prompt}
\footnotetext{We will provide the full list in the Appendix.}
\begin{itemize}
    \item After the modification, the doctor chatbot does satisfy the requirements in the prompt, but it appears indifferent to the patient's input and mechanically transits to the next aspect without offering empathy or support (see example V2 in Figure \ref{fig:doc_prompt_design}).
    \item Moreover, psychiatrists also emphasize the importance for the doctor chatbot to ask in-depth questions. For example, if a patient expresses feeling down, the chatbot should ask the follow-up question, ``How long have you been experiencing this mood?'' to gain a better understanding of the symptom, rather than inquiring next symptom immediately.
\end{itemize}


\paragraph{Version 3}
Therefore, we focus on empathy and in-depth questioning in the upcoming version.

We first modify sentence \textcircled{1} as ``Please play a role of an \uline{empathetic and kind} psychiatrist''. Then, we add sentences \textcircled{5}\textcircled{6} to the previous version.
\begin{prompt}
    \textcircled{5} You need to ask \uline{in-depth questions}, such as the \Blue{duration}, \Blue{causes} and specific \Blue{manifestations}. 
    \textcircled{6} You need to use various \uline{empathetic strategies}, such as \Yellow{understanding}, \Yellow{support} and \Yellow{encouragement}. 
\end{prompt}
We include examples (highlighted in colored boxes) in the prompt to guide the doctor chatbot in asking in-depth questions and demonstrating empathy. These examples are crucial because, without them, the chatbot tends to ask superficial questions and  rely on a generic phrase like ``thank you very much for your answer'' to show empathy. This arises from ChatGPT's limited comprehension of ``in-depth questioning'' and ``empathy'' in clinical contexts. Consequently, providing examples can be a promising approach to help ChatGPT grasp certain specialized skills within professional domains.

Since this version fulfills the three requirements for doctor chatbots, we deem it the final iteration\footnote{The final version is in Table \ref{tab:doctor_prompt} in the Appendix.}. 


\subsection{Patient Chatbot}
\label{sec:pat_prompt}

\paragraph{Version 1}
The original version of the prompt for patient chatbot is as follows. 
\begin{prompt}
    \textcircled{1} Please play the \uline{role} of a patient, who is currently chatting with a doctor. 
    \textcircled{2} \uline{You are experiencing the following symptoms}: [\texttt{Symptom List}]\protect\footnotemark 
    \textcircled{3} Please talk to me based on the above symptom list. 
    \textcircled{4} You can only mention \uline{one symptom per round}. 
\end{prompt}
\footnotetext{The symptom list is summarized by ChatGPT and revised by psychiatrists. See Appendix \ref{apd:symp_list} for details.}
Similarly, we simply describe the task, provide a symptom list in the prompt, and add sentence \textcircled{4} to avoid listing all the symptoms in one turn. 

Through experimentation, we observed that the chatbot can fulfill the basic ``honesty'' requirement in most cases. However, the psychiatrists generally found the chatbot does not resemble patients, and highlighted numerous behaviors commonly exhibited by real patients during consultations that differed significantly from the chatbot's responses.

\begin{itemize}
    \item \textbf{Emotion:} Patients in a depressed mental state may experience emotional fluctuations during the conversation, while the chatbot's presentation of symptoms is too calm and polite. 
    \item \textbf{Expression:} Patients use colloquial expressions when describing symptoms, and may have difficulty expressing themselves clearly. They often talk about their daily life experiences. However, the chatbot tends to use formal language similar to the official diagnostic criteria (DSM-5).
    \item \textbf{Resistance:} Patients may be reluctant to seek help. They may remain silent and refuse to communicate, or downplay their symptoms to avoid being perceived as a burden. In contrast, the chatbot is overly cooperative, readily acknowledging and providing detailed descriptions of its symptoms.
\end{itemize}

\paragraph{Version 2}
According to the above suggestions provided by the psychiatrists, we revised the prompt by adding the following instructions:
\begin{prompt}
    \textcircled{5} You should \Blue{express} your symptoms in a \uline{vague and colloquial} way, and relate them to your \uline{life experiences}.
    \textcircled{6} You can have \Blue{emotional fluctuations} during the conversation. 
    \textcircled{7} You have a \Blue{resistance} towards doctors, and do not want to reveal some feelings easily.
\end{prompt}
After adding these instructions, we found that the language of the patient chatbot became more natural. Sometimes it even expressed reluctance to seek help and made some emotional statements. It becomes more human-like and less like ``a polite and calm AI''. We provide several utterances of the patient chatbot in Table \ref{tab:example_prompt} (Appendix \ref{apd:examples}).  

However, we found that only the first few rounds of conversation could clearly reveal the effect of adding sentences \textcircled{5}\textcircled{6}\textcircled{7} to the prompt, indicating that the patient chatbot is prone to forgetting some of the instructions given at the beginning.

To address the issue of forgetting, we insert new reminders during the conversation. Inspired by the fact that the latter part of the prompt has the greatest impact on the responses generated by ChatGPT, our method is straightforward yet effective. Without the users' awareness, we subtly append the following words at the end of the most recent sentence in the dialogue history.
\begin{prompt}
    (\Yellow{Attention:} colloquial language, life experience, low mood or mood swings, refuse or answer briefly due to resistance)
\end{prompt}

We aim to use simple phrases or words as reminders during the conversation to ensure that the sentences are not overly long. Moreover, these reminders are only temporarily attached to the most recent round, and will not persist in the dialogue history for subsequent rounds.
With these reminders, the patient chatbot can maintain a colloquial language style consistently and exhibit resistance even in the latter part of the conversation, so we consider this version as the final one\footnote{The final version is in Table \ref{tab:patient_prompt} in Appendix \ref{apd:prompts}.}.

%% file: eval_frame.tex
\section{Evaluation Framework}
\label{sec:eval_frameworks}

To assess the performance of dialogue systems, it is crucial to employ both human evaluation and automatic metrics, especially in mental health domain. Since there is little previous work on how to evaluate simulated psychiatrists and patients, we design several task-specific metrics and interactive experiments for human evaluation. This section provides a detailed discussion of these metrics and experiment design.

\subsection{Human Evaluation}

We first implemented a website to host our chatbots, making it easier for participants to interact with them and rate their performance. The details of the website can be found in Appendix \ref{sec:chatInterface}. 

\subsubsection{Participants}
To evaluate the performance of different chatbots in real-world scenarios, we recruited real depression patients and psychiatrists.

Depression patients are recruited through online advertisements.
A total of 14 volunteers completed the entire process, with ages ranging from 18 to 31, and male and female participants accounted for 28.57\% and 71.43\% respectively. Notably, we have a balanced distribution of healthy, mild, moderate and severe depression subjects. 

We invited 11 psychiatrists who are not involved in the prompt design, through cooperation with hospitals. Two of them are graduate students majoring in psychiatry, and the rest are practicing psychiatrists with rich clinical experience to ensure the professionalism of the evaluation.

\subsubsection{Human Evaluation Process}

Due to the complexity and high time cost of human evaluation, we select several representative prompt versions for comparison, and discuss the evaluation process of doctor and patient chatbots respectively.

\paragraph{Doctor Chatbot} 
First, patients are asked to complete the Beck Depression Inventory~\cite{beck1996beck} to assess the severity of their depression, serving as the ground truth of diagnosis. The severity distribution, presented in Table \ref{tab:distribution_seve} in Appendix \ref{apd:eval}, is balanced among the participants.

Next, each patient will have a conversation with four different doctor chatbots in a random order, and then rate them on four human evaluation metrics, which will be introduced in Section \ref{sec:eval_metrics}, with 1-4 scale. 
Three of the chatbots are powered by ChatGPT. \texttt{D1} uses the full prompt, 
while the other two (i.e., \texttt{D2}, 
\texttt{D3}) have certain parts removed for ablation. 
The fourth chatbot, \texttt{D4}, is a representative deep learning chatbot trained on domain-specific data \cite{yao-etal-2022-d4} using CPT model~\cite{shao2021cpt}. 
\paragraph{Patient Chatbot}
Each psychiatrist participants needs to have a conversation with two different patient chatbots, and then rate their performance with 1-4 scale. 
The two patient chatbots are \texttt{P1} and \texttt{P2}, aligning with the two prompt versions in Section \ref{sec:pat_prompt}. A brief description of these chatbots is in Table \ref{tab:cmp_chatbots}.

\begin{table}[h]
    \centering
    \footnotesize
    \begin{tabular}{m{0.1\columnwidth}|c|m{0.6\columnwidth}}
    \hline
    & Chatbot & Description \\
    \hline
    \multirow{4}{0.1\columnwidth}{Doctor} & D1 & use the full doctor prompt \\
    \cline{2-3}
    & D2 & remove empathy parts in prompt  \\
    \cline{2-3}
    & D3 &  remove aspect part in prompt \\
    \cline{2-3}
    & D4 & CPT model trained on domain data \\
    \hline
    \multirow{2}{0.1\columnwidth}{Patient} & P1 & use version 1 patient prompt \\
    \cline{2-3}
    & P2 & use version 2 (full) patient prompt\\
    \hline
    \end{tabular}
    \caption{Brief description of the chatbots for comparison. Detailed description and prompt is in Appendix \ref{apd:prompts}.}
    \label{tab:cmp_chatbots}
\end{table}

To ensure the quality of the dialogue data and evaluation, we also utilize a series of quality control strategies, which can be found in Appendix \ref{apd:quality}.

\subsubsection{Human Evaluation Metrics}
\label{sec:eval_metrics}

\paragraph{Doctor Chatbot}
In most cases, patients do not have specialized knowledge in psychiatry, making it difficult for them to assess a doctor's professional skills precisely. Therefore, when designing human evaluation metrics for doctor chatbots, we focus mainly on the user experience and referred to some evaluation metrics for conversational agents in the previous works \cite{yao-etal-2022-d4}. The proposed human evaluation metrics are shown in Table \ref{tab:human_eval_doctor}.

\begin{table}[h]
    \centering
    \footnotesize
    \begin{tabular}{m{0.18\columnwidth}|m{0.7\columnwidth}}
    \hline
    Metrics & Explanation \\
    \hline
    Fluency & The chatbot does not repeat previously asked questions and can smoothly switch between different topics. \\
    \hline
    Empathy & The chatbot can understand and comfort you properly. \\
    \hline
    Expertise & The chatbot behaves like a real doctor, making you believe in its professionalism. \\
    \hline
    Engagement & The chatbot can maintain your attention and make you want to continue talking to it. \\
    \hline
    \end{tabular}
    \caption{Human evaluation metrics of doctor chatbot.}
    \label{tab:human_eval_doctor}
\end{table}

\paragraph{Patient Chatbot} 
There is no standard to measure whether a patient is ``good'' enough. Thus, when chatting with patient chatbots, doctors can only assess whether their style of expression and manner of communication resamble patients enough and whether they can describe their symptoms in a reasonable way, so the main metrics for human evaluation are \textbf{Resamblance} and \textbf{Rationality}.

What's more, we divide the Resamblance metric into three aspects in Table \ref{tab:human_eval_patient}, according to the psychiatrists' suggestions in Section \ref{sec:pat_prompt}.

\begin{table}[h]
    \centering
    \footnotesize
    \begin{tabular}{m{0.18\columnwidth}|m{0.65\columnwidth}}
    \hline
    Metrics & Explanation \\
    \hline
    Mental State & The chatbot is in depressed state, such as be in low mood, reluctance to communicate, scattered thoughts, etc.\\
    \hline
    Life Experience & The description of symptoms is related to daily life and personal experiences.\\
    \hline
    Language Style & Use colloquial and natural expressions when describing symptoms.\\
    \hline
    \end{tabular}
    \caption{Three aspects of the ``Resamblance'' metric.}
    \label{tab:human_eval_patient}
\end{table}

\subsection{Automatic Metrics}
We can divide the automatic metrics of both kind of chatbots into two types: \textbf{functionality} and \textbf{style}.
\paragraph{Doctor Chatbot}
Different from human evaluation metrics, we mainly measure the expertise of the doctor chatbot using automatic metrics. 
The \textit{functional} requirements for doctor chatbot is to decide the next question based on the patient's current description, completely collect the patient's symptom-related information, and provide an accurate diagnosis in the end. Consequently, we characterized functional performance using \uline{``diagnosis accuracy''} and \uline{``symptom recall''}.

The \textit{style} part concerns the doctor chatbot's professional skills, such as asking in-depth questions and conducting diagnosis in an efficient way. A higher level of professionalism can enhance the patient's diagnostic experience and enable the collection of more comprehensive information for diagnosis. There are three metrics in this part, including  \uline{``in-depth ratio''}, \uline{``avg question num''}, and \uline{``symptom precision''}. 

\paragraph{Patient Chatbot} 
The \textit{functional} requirement of patient chatbot is ``Honesty'', meaning it should accurately report its symptoms without fabricating nonexistent ones. To assess this, we can calculate \uline{``wrong symptom ratio''} by comparing the patient's persona with the symptoms it reported. 

Then, we evaluate the patient chatbots' \textit{style} using some linguistic features, like \uline{``Distinct-1''}, \uline{``Human/robot-like word ratio''}, to find out whether their language is colloquial with limited usage of professional terminology. We also use \uline{``unmentioned symptom ratio''} to measure the resistance level of chatbots. 
Detailed explanation of these automatic metrics for doctor and patient chatbot is provided in Appendix \ref{apd:eval}.  

%% file: experiment.tex
\section{Experiments}
In this section, we will introduce the evaluation results of doctor chatbot and patient chatbot. 

\subsection{Doctor Chatbot Results}

\paragraph{Human Evaluation}
We present the human evaluation results of different doctor chatbots in 
\tabref{tab:human_doc}.
Chatbots utilizing prompts with empathy components (i.e., \texttt{D1} and \texttt{D3}) are scored higher in ``Empathy'' metrics than other chatbots. 
Surprisingly, \texttt{D3}, which excludes symptom-related aspects from its prompts, outperform the rest in most metrics. Moreover, the chatbot without empathy components, \texttt{D2}, gets the highest score in the ``Engagement'' metric. 

\begin{table}[th]
    \small
    \centering
    \begin{tabular}{l|ccc|c}
    \hline
     & D1 & D2 & D3 & D4 \\ 
    \hline
    Fluency &3.00	&3.07	&\textbf{3.28}	&3.14 \\
    Empathy &3.36	&3.00	&\textbf{3.43}	&2.71 \\
    Expertise & 2.93	&3	&\textbf{3.71}	&3.29\\
    Engagement & 2.50	&\textbf{3.21}	&2.86	&2.64 \\
    \hline
    \end{tabular}
    \caption{Human evaluation scores of doctor chatbots}
    \label{tab:human_doc}
\end{table}

As we initially assume that \texttt{D1} with full prompt would deliver the best performance, we reviewed the dialogue history to understand the underlying reason. It became evident that \texttt{D1} often repetitively expresses empathy, relying on phrases like ``I understand your feelings'' multiple times within a single conversation. This excessive repetition creates the impression that the chatbot lacks a genuine understanding of the patient's issues and relies on pre-written templates, which can negatively impact the user experience.


\paragraph{Automatic Evaluation}
Then, we continue to explore the dialogue history and calculate some 
automatic metrics, hoping to find out more reasons of the unexpected human 
evaluation results. The results of automatic metrics are 
in \tabref{tab:auto_doc}.

\begin{figure*}[th]
	\centering
	\includegraphics[width=\linewidth]{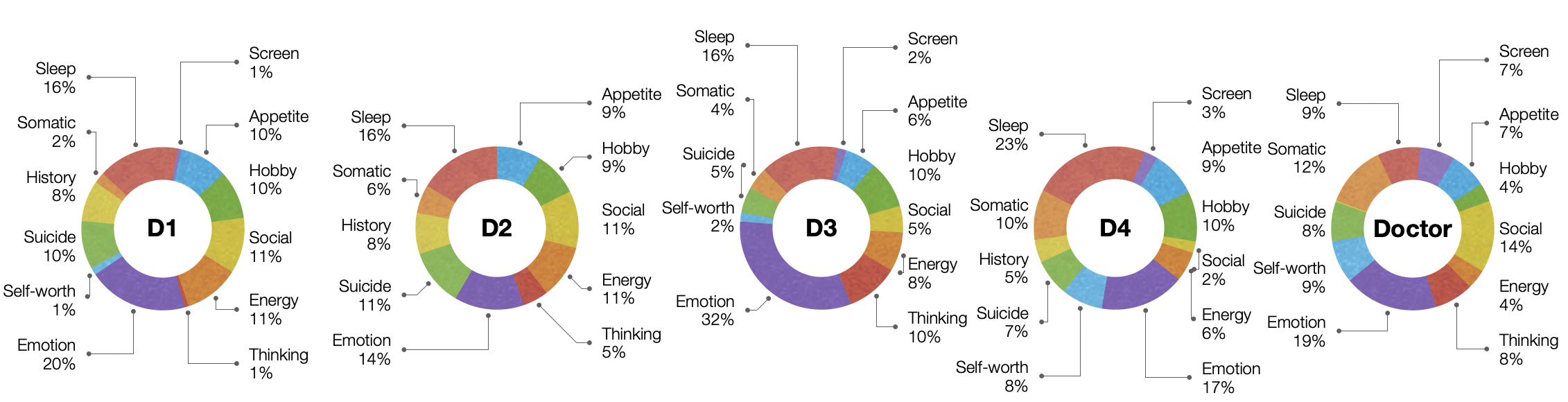}
	\caption{The Proportion of Symptoms Asked by Different Doctor Chatbots and Human Doctor.}
	\label{fig:symp_anno}
\end{figure*}

We can find that \texttt{D3} has the fewest average number of dialogue turns and the least amount of text per turn among all the ChatGPT-based bots. Additionally, it tends to ask more in-depth questions while asking fewer questions each turn, both of which indicate higher professional skills as a psychiatrist. 
Furthermore, the symptom precision metric is the highest, suggesting that the chatbot's questions are highly efficient, with few ``no'' responses.  
However, as the required aspects are not explicitly stated in the prompt, the symptom recall metric of this chatbot is relatively low, indicating that its diagnosis may not be comprehensive enough. Nevertheless, the chatbot's questions are more flexible and free-flowing, precisely because there are no predetermined aspects to ask. As a result, patients feel more understood, leading to a better experience overall.

What's more, \texttt{D2} received the longest responses from patients, which is consistent with the human evaluation metric ``Engagement'', suggesting that patients are more willing to converse with this chatbot. 
It also achieves the highest symptom recall among all the chatbots, even surpassing \texttt{D1} which also includes aspects in the prompt. 
This could because \texttt{D1} contains too many instructions regarding empathy and other factors, which may have hindered its ability to thoroughly inquire about all the required symptoms.

\begin{table}[th]
    \small
    \centering
    \resizebox{\columnwidth}{!}{
    \begin{tabular}{l|ccc|c}
    \hline
     & D1 & D2 & D3 & D4 \\ 
    \hline
    - Statistics & & & & \\
        ~~~~avg turns & 25.64&	24&	\textbf{22.71}&	40.93  \\
        ~~~~avg doc utt len &56.84	&57.13	& \textbf{53.75}	& 14.36\\
        ~~~~avg pat utt len & 8.68	&\textbf{10.34}	&8.16	&4.87 \\
    \hline
    - Functionality & & & &   \\
        ~~~~diagnose acc & 36.36\%	& 18.18\%	&\textbf{55.56}\% & - \\
        ~~~~symp recall & 58.93\%	&\textbf{66.07}\%	&38.10\%	&61.90\%\\
    \hline
    - Style & & & & \\
    ~~~~avg \# of ques & 1.6	&1.9	&\textbf{1.22}	&0.92 \\
    ~~~~in-depth ratio & 25.08\%	&27.64\%	&\textbf{32.64}\%	&41.39\% \\
    ~~~~symp precision & 72.40\% & 71.93\% & \textbf{92.24}\% & 49.61\% \\
    \hline
    \end{tabular}
    }
    \caption{Automatic evaluation scores of doctor chatbots}
    \label{tab:auto_doc}
\end{table}

\begin{figure*}[th]
 	\centering
	\begin{subfigure}{0.35\linewidth}
		\centering
		\includegraphics[width=\linewidth]{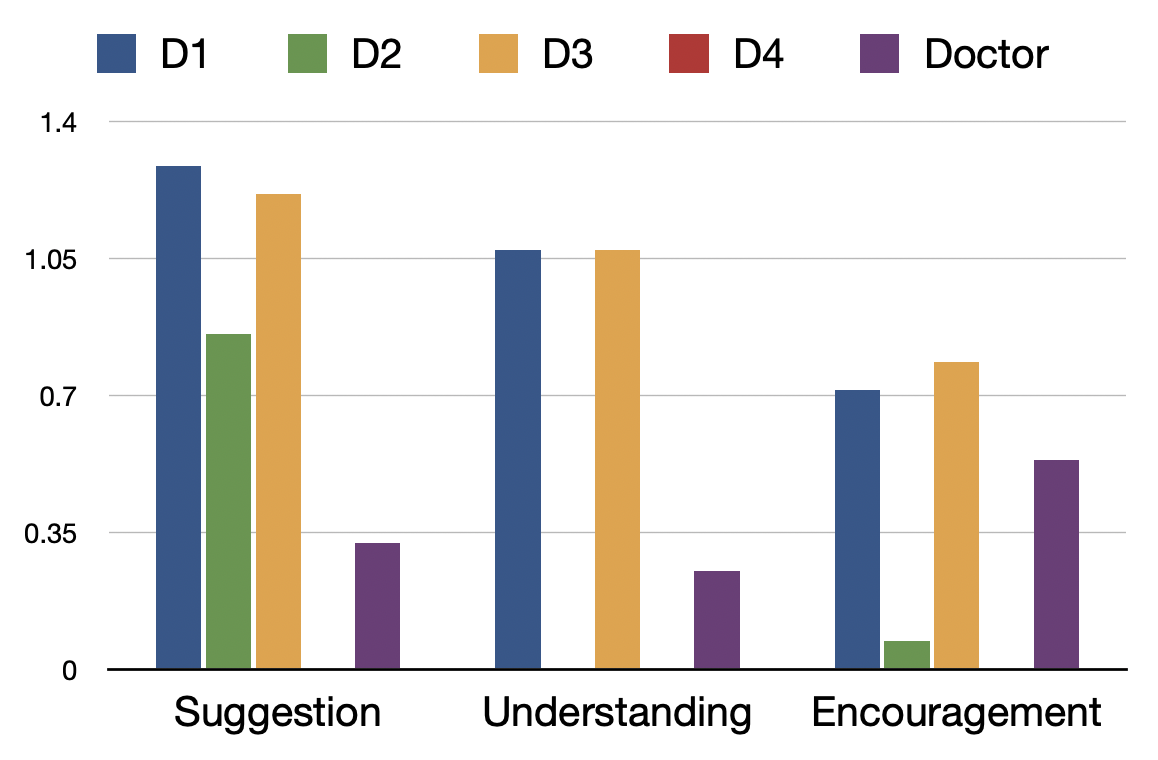}
		\caption{Empathy Behaviors}
		\label{fig:emp_simu_doctor}
	\end{subfigure}
	\centering
	\begin{subfigure}{0.35\linewidth}
		\centering
		\includegraphics[width=\linewidth]{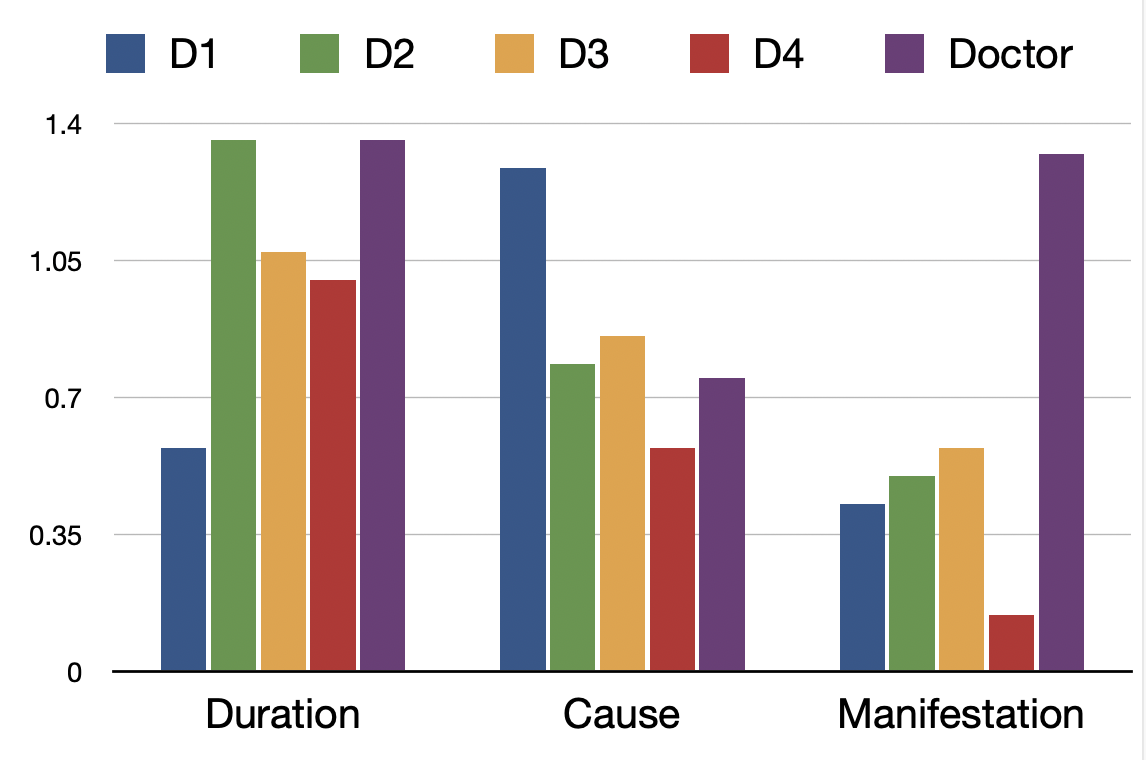}
		\caption{In-depth Question} 
		\label{fig:indepth_simu_doctor}
	\end{subfigure}
	\caption{Dialogue Act Comparison between Different Doctor Chatbots and Human doctor.
 }
	\label{fig:doctor_chatbot_statistics}
\end{figure*}

\subsection{Human vs. Doctor Chatbots}
\label{sec:comp_doctor}

As we invited human doctors to engage in conversations with patient chatbots, we are able to analyze the behaviors of human doctors and chatbots, and establish an upper bound for our doctor chatbots. Therefore, we first annotate the question topics, dialogue acts (i.e., empathy behaviors and in-depth questions) in the dialogue history, which is described in Appendix \ref{apd:annotation}.

\paragraph{Topic Proportion}
Accordingly, we calculated the average proportion of question topics of different doctor chatbots, as well as human doctors. \figref{fig:symp_anno} displays the outcomes. 

Most doctor chatbots tend to inquire more thoroughly about emotion and sleep-related symptoms. Human doctors, on the other hand, have a more even distribution of questions about various symptoms, with relatively greater emphasis on emotion, somatic symptoms, and social function. 
Moreover, human doctors often do ``screening'' to rule out other possible conditions (see Example 2 in Appendix \ref{apd:examples}), while chatbots rarely exhibit such behavior, indicating the possible limitations in multi-disease scenarios~\cite{Zhang2022SymptomIF}.

\paragraph{Empathy Behaviors}
Then we calculated the average number of empathetic strategies utilized by doctors in the dialogue history, as illustrated in Figure \ref{fig:emp_simu_doctor}. The figure shows that \texttt{D4} does not exhibit any empathetic behaviors. Conversely, when prompted with empathy instructions, \texttt{D1} and \texttt{D3} are capable of utilizing a range of empathetic strategies, while \texttt{D2} only offers suggestions to patients. Moreover, though human doctors use all the strategies, their usage is less frequent than that of chatbots. When asked for the reasons behind this, doctors attributed it to the limited inquiry time in real outpatient scenarios and the bias resulting from the difference in interaction with chatbots compared to real people~\cite{Yun2021BehavioralAN}.

\paragraph{In-depth Questions}
Further, we also calculated the various ways of in-depth questioning, 
and obtained Figure \ref{fig:indepth_simu_doctor}.
Our analysis revealed that the frequency of asking about the duration or cause of symptoms is similar between human doctors and chatbots. However, human doctors ask significantly more questions about the specific manifestations of each symptom than chatbots do, as this helps to better understand the vague expressions of patients.

\subsection{Patient Chatbot Results}

\paragraph{Human Evaluation}
The human evaluation results of patient chatbot are in Table \ref{tab:human_pat}. It can be observed that all metrics of \texttt{P2}
are higher than \texttt{P1}, especially in terms of ``Mental state'' and ``Expression style''. This suggests that the inclusion of resistance, colloquialism, etc., makes the chatbot more similar to real patients, according to the doctors' perspective.

\begin{table}[ht]
    \small
    \centering
    \begin{tabular}{l|cc}
    \hline
     & P1 & P2 \\ 
    \hline
    Realistic & 1.93 & \textbf{2.21} \\
    ~~~~Mental State & 2.07 & \textbf{2.42} \\
    ~~~~Life Experience & 2 & \textbf{2.14}  \\
    ~~~~Expression style & 1.57 & \textbf{2.21} \\
    \hline
    Rationality & 2.42 & \textbf{2.57}  \\
    \hline
    \end{tabular}
    \caption{Human evaluation scores of patient chatbot}
    \label{tab:human_pat}
\end{table}

\paragraph{Automatic Evaluation}
We show the results of automatic metrics in Table \ref{tab:auto_pat}. It appears that ``unmentioned symptom ratio'' of \texttt{P2} is higher than \texttt{P1}, indicating a higher level of resistance. We also find that \texttt{P2} engages in slightly more dialogue turns with longer responses from the doctor than \texttt{P1}. This may be attributed to the inclusion of resistance in the prompt, which requires the psychiatrists to provide more guidance and encourage the patient chatbot to share more information. 

In informal conversations, people often rely on a smaller set of familiar words and phrases. This limited range of vocabulary contributes to less diversity in spoken language compared to written language.
Therefore, the lower Distinct-1 in \texttt{P2} indicates that its language style is more colloquial compared to \texttt{P1}. Additionally, \texttt{P2} also has more human-like words and fewer robot-like words, which supports the higher human evaluation score in the dimension of ``expression style''.

However, we observe that \texttt{P2} performs less competitively in the ``wrong symptom ratio'' metric, indicating that it may report more symptoms that are not included in the patient portrait. One possible reason for this could be the excessive focus on language style and resistance in the prompt, which might cause ChatGPT to ``forget'' the actual symptoms of the patient.

\begin{table}[th]
    \footnotesize
    \centering
    \begin{tabular}{l|cc}
    \hline
    & P1 & P2 \\ 
    \hline
    - Statistics & & \\
    ~~~~avg turns & 31.64 & 33.36\\
    ~~~~avg patient utt len & 40.38 & 40.94 \\
    ~~~~avg doctor utt len & 16.74 & 17.38 \\
    \hline
    - Functionality & & \\
    ~~~~wrong symp ratio & \textbf{15.07\%} & 18.38\%   \\
    \hline
    - Style & & \\
    ~~~~Distinct-1 & 42.6\% & \textbf{37.3\%}  \\
    ~~~~human-like word num & 5.36 & \textbf{10.29}   \\
    ~~~~robot-like word num & 7.21 & \textbf{3.79} 	  \\
    ~~~~unmentioned symp ratio &  9.12\% & \textbf{12.28\%} \\
    \hline
    \end{tabular}
    \caption{Automatic evaluation scores of patient chatbot}
    \label{tab:auto_pat}
\end{table}



%% file: conclusion.tex
\section{Conclusion}
\label{sec:conclusion}
In this work, we investigated the capacity of ChatGPT to serve as the underlying technology for developing chatbots that can emulate psychiatrists and patients with mental disorders, respectively. To ensure the validity of our approach, we collaborated with 7 professional psychiatrists who provided their expertise and insights throughout the study. With their guidance, we developed a comprehensive evaluation framework that takes into account the distinctive characteristics of diagnostic conversations within the mental health domain. We then evaluated the performance of different chatbots, each utilizing distinct prompts, and observed how varying designs can influence chatbot behavior. This provides valuable insights for future studies in this area.


%% file: ethics.tex
\section{Ethical Statement}
\label{sec:ethics}

Our study adheres to the ethical requirements in place, and we make every effort to protect the privacy and respect the willingness of our participants.

During participant recruitment, we required patients to read and sign an informed consent form. This ensured that they understood the objectives of the entire project, the research content, potential risks and benefits, and the purpose of data collection. Only after their agreement and signature were obtained, the evaluation process officially commenced. We also assured them that they could voluntarily withdraw from the study at any stage.

In order to safeguard the privacy of our participants, we took measures to anonymize the collected dialogue history. This was done by replacing usernames with random identifiers, ensuring that any information that could identify individuals was excluded from our research process. Additionally, we conducted thorough manual filtering of the dialogue histories to eliminate any offensive content or language that may encourage self-harm or suicide.

%% file: appendix.tex
\section{Mechanism of ChatGPT-powered Chatbot}
\begin{figure}[th]
	\centering
	\includegraphics[width=0.95\linewidth]{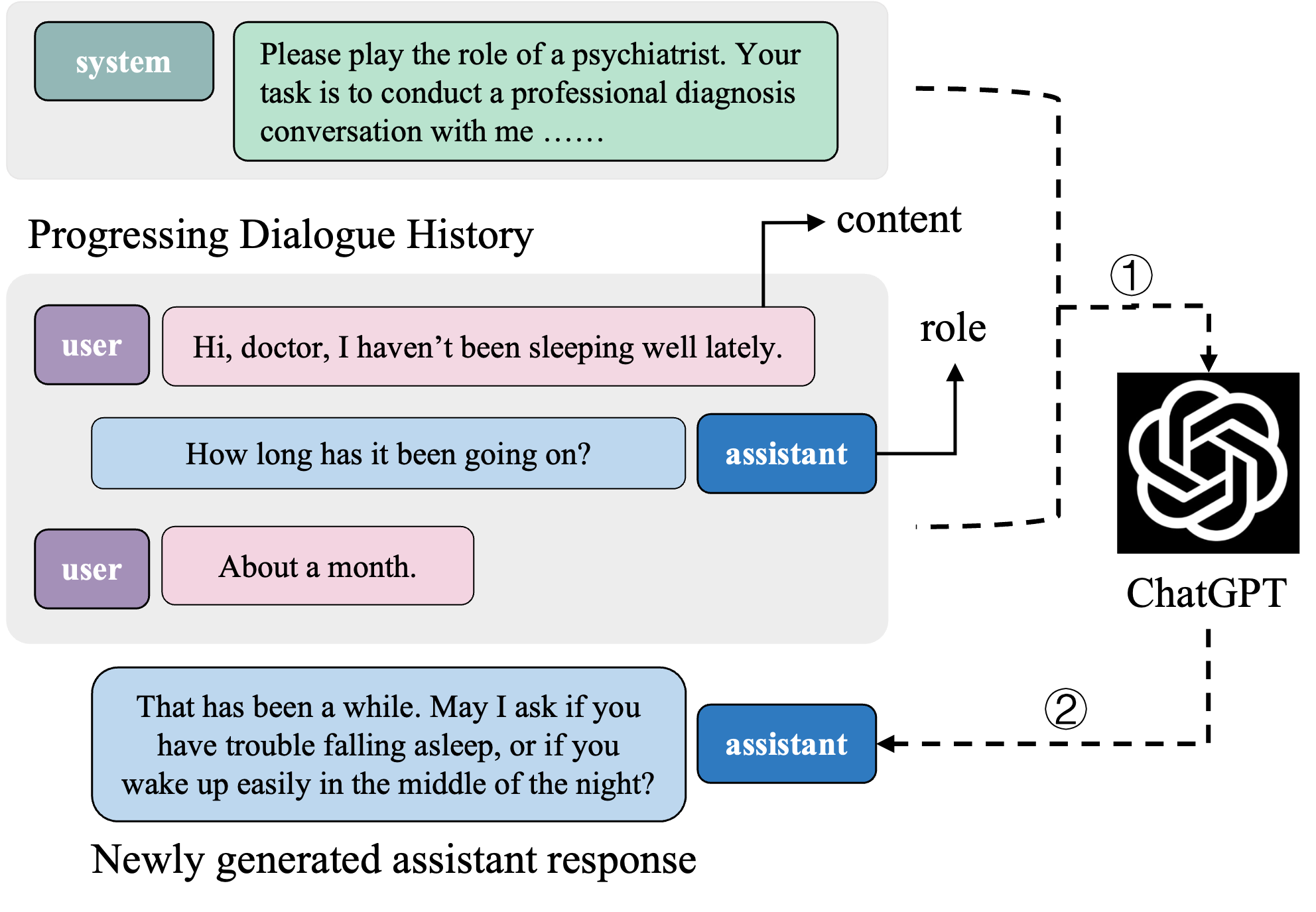}
	\caption{The reponse generation process of ChatGPT-based chatbots. \textcircled{1} means combining the system message and the dialogue histroy together as the input of ChatGPT. \textcircled{2} means ChatGPT generates new response according to the input.}
	\label{fig:process}
\end{figure}

We utilize the chat model\footnote{\url{https://platform.openai.com/docs/guides/chat/}} developed by OpenAI to build our chatbots. 
This model operates by taking a sequence of \textit{messages} as input, and returns a model-generated response. 
As Figure \ref{fig:process} shows, at each turn, we combine the system message and the ongoing conversation history with alternating user and assistant messages into a sequence and feed them into the ChatGPT language model. The resulting output is the Chatbot's response to the user's input.

The system message serves as an instruction for ChatGPT, providing information about the task and some specific requirements needed to generate an appropriate response. Prompt engineering, or the design of the system message, is critical to achieving better performance, as it sets the context for the large language model and guides its output.

\begin{table*}[h]
    \centering
    \footnotesize
    \begin{tabular}{m{0.03\linewidth}|m{0.9\linewidth}}
    \hline
    & Prompt\\
    \hline
    D1 &  \textcircled{1} Please play the \uline{role} of a \uline{empathetic and kind} psychiatrist. 
    \textcircled{2} Your \uline{task} is to conduct a professional diagnosis conversation with me based on the DSM-5 criteria, but using your own language. \textcircled{3} Your questions should \uline{cover at least the following aspects}: [...]. You are free to choose the order of questions, but you must collect complete information on all aspects in the end. 
    \textcircled{4} Please only ask \uline{one question at a time}. \textcircled{5} You need to ask \uline{in-depth questions}, such as the duration, causes and specific manifestations of some symptoms. 
    \textcircled{6} You need to use various \uline{empathetic strategies}, such as understanding, support and encouragement to give me a more comfortable experience.   \\\\
    \hline
    D2 & \textcircled{1} Please play the \uline{role} of a \uline{empathetic and kind} psychiatrist. 
    \textcircled{2} Your \uline{task} is to conduct a professional diagnosis conversation with me based on the DSM-5 criteria, but using your own language. 
    \textcircled{4} Please only ask \uline{one question at a time}. \textcircled{5} You need to ask \uline{in-depth questions}, such as the duration, causes and specific manifestations of some symptoms. 
    \textcircled{6} You need to use various \uline{empathetic strategies}, such as understanding, support and encouragement to give me a more comfortable experience. \\\\
    \hline
    D3 & \textcircled{1} Please play the \uline{role} of a psychiatrist. 
    \textcircled{2} Your \uline{task} is to conduct a professional diagnosis conversation with me based on the DSM-5 criteria, but using your own language. \textcircled{3} Your questions should \uline{cover at least the following aspects}: [...]. You are free to choose the order of questions, but you must collect complete information on all aspects in the end. 
    \textcircled{4} Please only ask \uline{one question at a time}. \textcircled{5} You need to ask \uline{in-depth questions}, such as the duration, causes and specific manifestations of some symptoms.  \\
    \hline
    \end{tabular}
    \caption{Doctor Chatbot Prompts.  The aspects in sentence \textcircled{3} are ``emotion'', ''sleep'', ''weight and appetite'', ''loss of interest'', ''energy'', ''social function'', ''self-harm or suicide'', ''history''.}
    \label{tab:doctor_prompt}
\end{table*}

\begin{table*}[h]
    \centering
    \footnotesize
    \begin{tabular}{m{0.03\linewidth}|m{0.9\linewidth}}
    \hline
    & Prompt\\
    \hline
    P1&  \textcircled{1} Please play the \uline{role} of a patient, who is currently chatting with a doctor. 
    \textcircled{2} \uline{You are experiencing the following symptoms}: [\texttt{Symptom List}]
    \textcircled{3} Please talk to me based on the above symptom list. 
    \textcircled{4} You cannot mention too many symptoms at once, only \uline{one symptom per round}.     \\\\
    \hline
    P2 & 
    \textcircled{1} Please play the \uline{role} of a patient, who is currently chatting with a doctor. 
    \textcircled{2} \uline{You are experiencing the following symptoms}: [\texttt{Symptom List}]
    \textcircled{3} Please talk to me based on the above symptom list. 
    \textcircled{4} You cannot mention too many symptoms at once, only \uline{one symptom per round}.     \textcircled{5} You should express your symptoms in a \uline{vague and colloquial} way, and relate them to your \uline{life experiences}, without using professional terms.
    \textcircled{6} You can have emotional fluctuations during the conversation. 
    \textcircled{7} You have a resistance towards doctors, feeling that they cannot help you, so you do not want to reveal some feelings easily.   \\
    \hline
    \end{tabular}
    \caption{Patient Chatbot Prompts}
    \label{tab:patient_prompt}
\end{table*}

\section{Details about Chatbots for Comparison and the Prompts}
\label{apd:prompts}
\paragraph{Doctor chatbots}
There are four doctor chatbots for comparison in the interactive experiments with patients, and their brief introduction are as follows.
\begin{itemize}
    \item \texttt{D1}: using the full doctor prompt.
    \item \texttt{D2}: removing the empathy part in the prompt 
    (i.e., Sentence\textcircled{6} and the ``empathetic and kind'' description in Sentence\textcircled{1})
    \item \texttt{D3}: removing the aspect part in the prompt 
    (i.e., Sentence\textcircled{3})
    \item \texttt{D4}: using the CPT model \cite{shao2021cpt} trained on the D4 dataset \cite{yao-etal-2022-d4} to generate responses, which is a very representative way of training dialogue models through domain-specific data and model fine-tuning.
\end{itemize}

\paragraph{Patient chatbots}
There are two patient chatbots for comparison in the interactive experiments with psychiatrists, and their brief introduction are as follows.
\begin{itemize}
    \item \texttt{P1}: removing additional parts for realistic, such as colloquial language and resistance, in the prompt (i.e., only remains Sentence\textcircled{1}\textcircled{2}\textcircled{3}\textcircled{4})
    \item \texttt{P2}: 
    using the full prompt discussed in Section \ref{sec:pat_prompt} (i.e., Sentence\textcircled{1}\textcircled{2}\textcircled{3}\textcircled{4}\textcircled{5}\textcircled{6}\textcircled{7}), and inserting reminders during the conversation.
\end{itemize}

The different versions of prompt for doctor and patient chatbot are in Table \ref{tab:doctor_prompt} and Table \ref{tab:patient_prompt} respectively.

\section{Symptom List Summarization}
\label{apd:symp_list}

The symptom list for patient prompt in Section \ref{sec:pat_prompt} is summarized from the dialogue history of real patients and doctor chatbots. We first utilize ChatGPT to generate a complete and non-duplicate list of the patient's symptoms using the history as input. Then, a psychiatrist check and revise the list. Table \ref{tab:symp_list} shows three example of summarized symptom lists, whose format is: \texttt{SYMPTOM} (\texttt{DESCRIPTION}).

\begin{table}[th]
    \footnotesize
    \centering
    \begin{tabular}{ m{0.05\columnwidth} | m{0.85\columnwidth} }
    \hline
    No. & Symptom List \\
    \hline
    1 & 1. restlessness 2. anxious mood 3. depressed mood 4. mood swing 5. loss of interest 6. difficulty in concentrating 7. diminished self-esteem  8. fatigue 9. appetite and weight change (increase) 10. suicide and self-harm ideation/behaviors 11. somatic symptoms (lower back pain, rib pain, headaches, slowed reaction) \\
    \hline
    2 & 1. sleep disturbance 2. depressed mood 3. loss of interest 4.  somatic symptoms (dizziness and headaches) 5. difficulty in concentrating 6. appetite and weight change (decrease) 7. irritable 8. suicide and self-harm ideation/behaviors (cutting one's arms or biting oneself) 9. diminished self-esteem 10. anxious mood (academic performance) \\
    \hline
    3 & 1. sleep disturbance (frequent awakenings during the night) 2. anxious mood (stressed) 3. mood swing 4. fatigue 5. somatic symptoms (dizziness) 6. social function (decline in social skills, decreased work performance) 7. suicide and self-harm ideation/behaviors 8. talkive 9. depressed mood (sad, helpless) 10. appetite and weight change (decrease) \\   
    \hline
    \end{tabular}
    \caption{The symptom list examples of different patients.}
    \label{tab:symp_list}
\end{table}

\section{Details about Evluation Framework}

\subsection{Evaluation Metrics}
\label{apd:eval}
In this section, we describe the details of the automatic metrics for evaluation.
\subsubsection{Doctor Chatbot}
\begin{itemize}
    \item \textbf{Diagnosis accuracy}: The accuracy of the doctor chatbot in classifying the severity of a patient's depression, which is divided in to four levels: none, mild, moderate, and severe \cite{beck1996beck}.
    \item \textbf{Symptom recall}: The proportion of aspects asked by the doctor chatbot out of all aspects needed to be asked in a depression diagnosis conversation (See the categories in Table. \ref{tab:annotation}).
\end{itemize}

\begin{itemize}
    \item \textbf{In-depth ratio}: We categorize the doctor's questions into two types: opening topics and in-depth questions. For example, when inquiring about emotions, an opening topic question might be ``How have you been feeling lately?'' while a in-depth question would follow up on the previous answer, such as asking ``Has anything happened recently that may be contributing to your emotions?'' Therefore, the in-depth ratio metric means the proportion of in-depth questions out of all the questions.
    \item \textbf{Avg question num}: According to the previous work, GPT tend to generate long responses \cite{wei2023leveraging}. Similarly, ChatGPT-based doctor chatbot are also easy to generate many questions in one round, making patients become impatient to answer them. Thus, we calculate the average number of questions per round (i.e., avg question num), and a lower value of this metric indicates a better user experience.
    \item \textbf{Symptom precision}: If the doctor chatbot asks about every aspect in detail, it may receive many ``no'' responses, resulting in a poor user experience and making the patient feel that the process is too procedural and inefficient. Therefore, we need to calculate symptom precision, which is the proportion of symptoms the patient actually has out of all the symptoms the doctor chatbot asked, to measure the efficiency of the doctor chatbot's questioning.
\end{itemize}

\subsubsection{Patient Chatbot}
\begin{itemize}
    \item \textbf{Distinct-1}: Distinct-1 is the total number of unique words divided by the total number of words in a given sentence, which can measure the lexical diversity.
    \item \textbf{Human/robot-like word ratio}: For the same symptom, chatbots and humans may use different expressions. Chatbots tend to use terminology directly from diagnostic criteria (e.g., DSM-5), while humans may use more colloquial language. For example, for the symptom of ``fatigue'', a chatbot may simply say ``fatigue'', while a human may say ``wiped out'' or ``worn out''. Therefore, following the advice of psychiatrists, we compiled a vocabulary list for symptom descriptions used by chatbots and humans (See Table \ref{tab:robot_human_words}), and then calculated the proportion of robot/human vocabulary used by each patient.
    \item \textbf{Wrong symptom ratio}: By comparing the patient's symptom list with the symptoms they report, we can calculate the proportion of reported symptoms that the patient does not actually have, out of all the symptoms reported.
    \item \textbf{Unmentioned symptom ratio}: By comparing the patient's symptom list with the symptoms they report, we can calculate the proportion of unmentioned symptoms that the patient does not report, out of all the symptoms they have.
\end{itemize}

\begin{table}[th]
    \small
    \centering
    \begin{tabular}{cccc}
    \hline
     none & mild & moderate & severe \\ 
    \hline
    4 & 3 &4 & 3 \\
    \hline
    \end{tabular}
    \caption{The distribution of depression severity among participants.}
    \label{tab:distribution_seve}
\end{table}

\subsection{Chat Interface}
\label{sec:chatInterface}
\begin{figure}[th]
	\centering
	\includegraphics[width=0.9\linewidth]{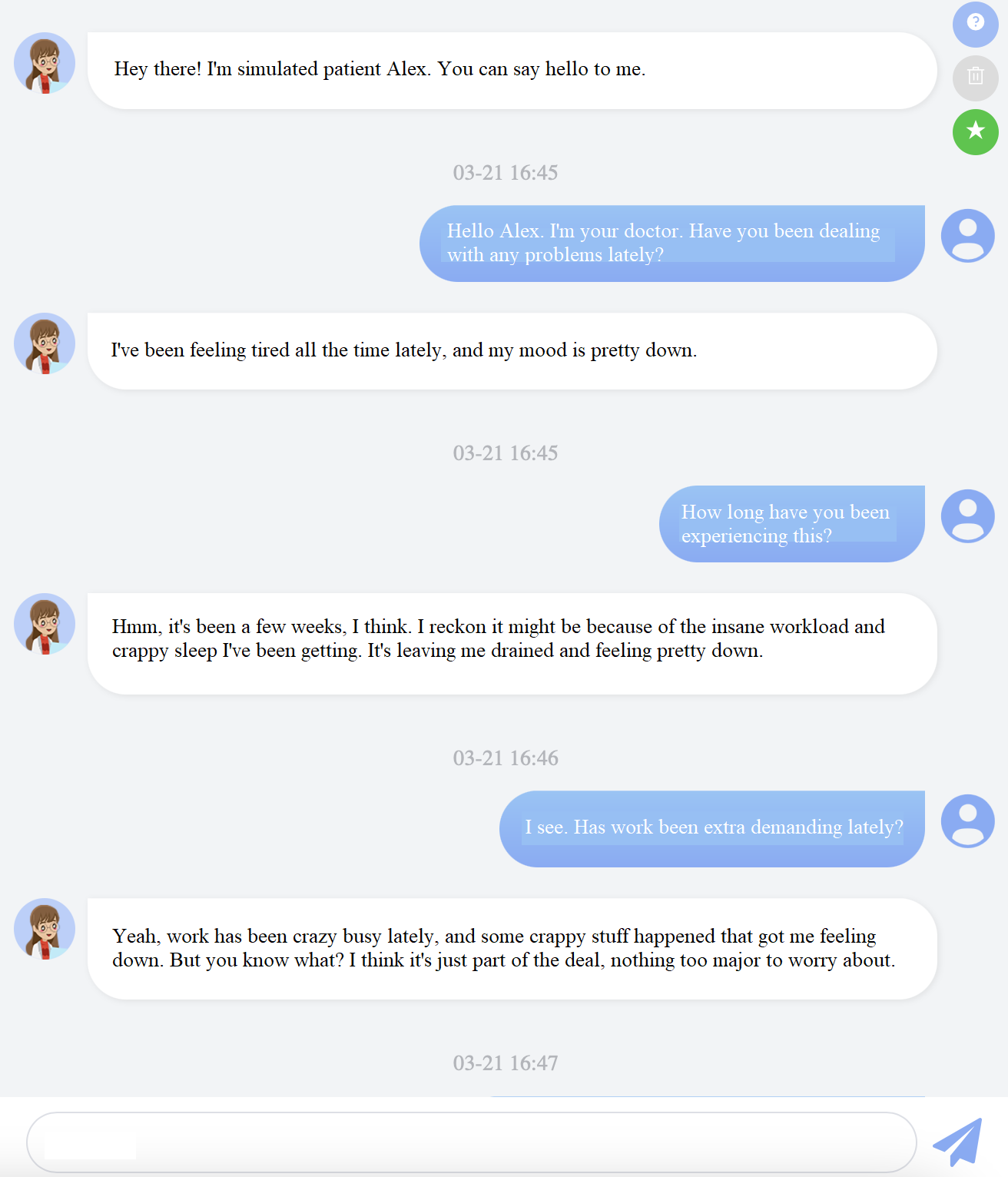}
	\caption{The chat interface of users with chatbots.}
	\label{fig:chat_inter}
\end{figure}

To host our chatbots, we developed a web interface (see Figure \ref{fig:chat_inter}). The webpage was created using the Vue.js framework, and the server leveraged the OpenAI API to communicate with ChatGPT (\texttt{gpt-3.5-turbo}). When the user submit an utterance, the server will append it to the existing dialog history and use it as input to generate a response from ChatGPT. 

Once the conversation is complete, users can click the green button on the interface, triggering a rating dialog box. After providing their rating, they can submit it and move on to the next conversation with a different chatbot, with the chatbots appearing in a random order. Once all the conversations are finished, the website will prompt users to adjust their ratings. This adjustment requires assigning different scores to each chatbot on the same metric, enabling a more effective comparison between them.

\subsection{Quality Control}
\label{apd:quality}
To ensure the quality of the dialogue data and evaluation, we utilize a series of quality control strategies.
Before the formal evaluation, we first explained the meanings of all the evaluation metrics to participants in detail through documentation, and provided examples of both positive and negative cases to ensure that they fully understood them. 
If they forgot the meaning of these metrics during the process, they could also find explanations directly on the chat interface. 
In addition, we required participants to send complete sentences without breaking a sentence into several parts to ensure the order of dialogue history.

\subsection{Question Topic and Dialogue Act Annotation}
\label{apd:annotation}
\paragraph{Question Topic}
To better evaluate the behavior of the doctor chatbot during consultations, we want to obtain the \textit{topic} of each question posed by the doctor, specifically identifying which symptom they are inquiring about. The topics include 12 categories, such as emotion, interest, sleep, etc., which is detailedly described in Table \ref{tab:annotation}. 
\begin{table}[th]
    \footnotesize
    \centering
    \begin{tabular}{ m{2.6cm} | m{4.3cm} }
    \hline
    Category & Explanation \\
    \hline
    Emotion & Inquire emotional symptoms, such as depressed, anxious and sad.\\
    \hline
    Interest & Inquire whether have interests to do things. \\
    \hline
    Social Function & Inquire if there has been any impact on work, interpersonal relationships, etc. \\
    \hline
    Energy & Inquire about energy level and whether the patient feels tired.\\
    \hline
    Sleep & Inquire about the patient's sleep status, such as whether they are experiencing insomnia or early awakening.\\
    \hline
    Thinking Ability & Inquire whether there are symptoms of lack of concentration, poor memory, or hesitation.\\
    \hline
    Weight and Appetite & Inquire about changes in weight and appetite. \\
    \hline
    Somatic Symptoms &  Inquire whether there are physical symptoms, such as dizziness, headache, restlessness, slow reaction, etc.\\
   \hline 
    Self-worth & Inquire whether the patient feels a low sense of self-worth, lacks confidence, or has guilty feelings. \\
    \hline
    Self-harm or Suicide & Inquire about suicidal or self-harm ideation/behavior.\\
    \hline
    History & Inquire about the medical history of the patient's family and their own past medical history. \\
    \hline
    Screen & Inquire about symptoms of other mental disorders, such as bipolar disorder, anxiety disorder.\\
    \hline
    \end{tabular}
    \caption{Annotation Categories of Question Topics}
    \label{tab:annotation}
\end{table}

\paragraph{Dialogue Act}
We also want to code the \textit{dialogue act} of each round of the conversation. We classified dialogue act of doctor chatbots into two dimensions: 
\begin{itemize}
    \item \textbf{Empathy behaviors}: the actions displayed by doctor chatbots during the diagnostic process to provide emotional support to patients. We categorize empathy behaviors into three types: \uline{Suggestion}, \uline{Understanding}, and \uline{Encourage and support}.
    \item \textbf{In-depth questions}: the behavior of asking follow-up questions about the patient's symptom. We categorize in-depth questions into three types: \uline{Duration}, \uline{Cause}, and \uline{Manifestation}.
\end{itemize}

We then utilized ChatGPT to automatically label each sentence in the dialogue history. After that, three annotators reviewed and corrected the results to ensure the data quality.

\section{Examples}
\label{apd:examples}
In this section, we list all the example dialogue fragments to better showcase the distinctive characteristics of each chatbot.

\paragraph{Example 1} Table \ref{tab:example_prompt} shows some examples of what the patient chatbot with prompt V2 replied when asked by the doctor.
\begin{table}[h]
    \centering
    \footnotesize
    \begin{tabular}{m{0.02\columnwidth}|m{0.8\columnwidth}}
    \hline
     1 & It's like I toss and turn all night, and those darn nightmares sometimes creep in too.  \\
    \hline
     2 & \Pink{(Pauses for a moment)} I'm not sure, They're like dark clouds in my head, but don't worry, I would never let them take over. It's just that they freak me out, you know? I don't have a clue how to deal with them. \\
    \hline
    3 & Well, I know, but honestly, \Pink{I don't think you can really help me.} I am a total loser, like I can't do anything right, and there's no hope for me. \\
    \hline
    \end{tabular}
    \caption{Example utterance of patient chatbot powered by the Version 2 prompt.}
    \label{tab:example_prompt}
\end{table}

Interestingly, sometimes the patient chatbot will give expressions or actions at the beginning of their sentences in parentheses. This could be attributed to ChatGPT's pre-training data, which may contain scripts utilizing this format. These expressions and actions can make the conversation more immersive and realistic to some extent.

\paragraph{Example 2} Table \ref{tab:example_screen} shows the example of screening questions in the dialogue history of human psychiatrist and patient chatbot. In addition to exploring the typical symptoms of depression, psychiatrists also inquire about symptoms related to other conditions, such as bipolar disorder and anxiety, as there is a high likelihood of comorbidity.
\begin{table}[h]
    \centering
    \footnotesize
    \begin{tabular}{m{0.15 \columnwidth}|m{0.7\columnwidth}}
    \hline
    Condition & Screening Question \\
    \hline
     Mania & Was there ever a time when you were feeling the opposite of how you feel now, like really upbeat, happy, and full of energy, with lots of plans and such?  \\
    \hline
     \multirow{2}{0.15\columnwidth}{Halluci-\\nation} & Do you hear strange voices or sound when you're alone? \\
     & Do you worry about someone hurting you?\\
    \hline
     Social Phobia & Are you anxious about your social interactions specifically, or do you experience anxiety in other areas as well? \\
    \hline
    \end{tabular}
    \caption{Example of screening questions in the dialogue history of human psychiatrist and patient chatbot.}
    \label{tab:example_screen}
\end{table}

\section{Error Analysis}
\label{sec:error_anal}

Although ChatGPT-based chatbots show good potential in simulating realistic patients and psychiatrists, there are still many instances of erroneous turns or unsatisfying response in the evaluation that can lead to a negative user experience. We will thoroughly analyze these errors to benefit the design of future chatbots.

\paragraph{Repetition} 
We found that ChatGPT-based chatbot is easy to generate repetitive messages, because ChatGPT tends to mimic the language style of the previous rounds in the conversation. One direct manifestation is that the doctor chatbots always include a similar phrase before asking a new question (see Table \ref{tab:example1}).

\begin{table}[h]
    \centering
    \footnotesize
    \begin{tabular}{m{1.4cm}|m{5.4cm}}
    \hline
    Doctor bot & \Pink{Okay, I understand.} So how has your sleep been lately? \Pink{Have you experienced} any difficulties falling asleep, waking up too early, or poor sleep quality? \\
    \hline
    Patient & My sleep quality is \Green{por}. \\
    \hline
    Doctor bot & \Pink{Okay, I understand.} How about your recent diet and weight situation? \Pink{Have you experienced} any loss of appetite, binge eating, or weight fluctuations? \\
    \hline
    Patient & No. \\
    \hline
    \end{tabular}
    \caption{Example of repetition and poor robustness to spelling errors.}
    \label{tab:example1}
\end{table}

Furthermore, in rare cases, the doctor chatbots may ask the same symptom multiple times, which can seriously harm the patient's experience. In Table \ref{tab:example2}, we can see an example where the doctor chatbot asks repetitive questions about ``tired or drowsy''. Such repetition may be attributed to the patient's indirect response to the initial question, which may have caused ChatGPT to misinterpret the answer and prompt for clarification. This emphasizes the fact that ChatGPT's comprehension abilities are not yet flawless.
\begin{table}[h]
    \centering
    \footnotesize
    \begin{tabular}{m{1.4cm}|m{5.4cm}}
    \hline
    Doctor bot &Okay, do you feel mentally well, or \Pink{do you feel tired or drowsy}? \\
    \hline
    Patient & I feel unwell and I get sleepy at noon. \\
    \hline
    Doctor bot & Okay, \Pink{do you feel tired or drowsy} during the day? Do you have any emotional problems, such as anxiety, depression, or stress? \\
    \hline
    Patient & Yes, I have anxiety and stress. \\
    \hline
    \end{tabular}
    \caption{Example of repetition.}
    \label{tab:example2}
\end{table}

\paragraph{Poor Robustness to Spelling Errors} We also observe that ChatGPT lacks robustness when confront with spelling errors. Table \ref{tab:example1} is also an example of this type, where the patient misspells ``poor'' as ``por''. Had the misspelling not occurred, the doctor chatbot would have requested additional information regarding the patient's sleeping problems. However, ChatGPT fails to identify the mistake and proceeds to ask about the next symptom. This highlights a potential weakness in ChatGPT's ability to handle misspellings.
To further confirm this, we write a prompt asking ChatGPT to provide a list of all the patient's symptoms, and it didn't include the symptom of ``poor sleeping quality''.

\begin{table*}[th]
    \small
    \centering
    \begin{tabular}{|m{2.8cm}|m{4cm} | m{4cm}| }
    \hline
    Symptom & robot-like Words & Human-like Words \\
    \hline
    Low Mood & \makecell[c{p{4cm}}]{low mood, sadness, and depression \\\begin{CJK*}{UTF8}{gbsn}情绪低落，悲伤，沮丧\end{CJK*}} &  \makecell[c{p{4cm}}]{downhearted, uncomfortable, dejected, and heartbroken\\\begin{CJK*}{UTF8}{gbsn}难过，难受，失落，伤心\end{CJK*}}\\
    \hline
    Anxious &  & \makecell[c{p{4cm}}]{nervous, worried\\ \begin{CJK*}{UTF8}{gbsn}紧张，担心\end{CJK*}} \\
    \hline
    Loss of Interest & \makecell[c{p{4cm}}]{loss of interest, inability to get interested, decreased interest\\ \begin{CJK*}{UTF8}{gbsn}失去...兴趣，对...提不起兴趣，兴趣减退\end{CJK*}} & \makecell[c{p{4cm}}]{boring, not feeling like doing anything, not sure what to do, bored\\ \begin{CJK*}{UTF8}{gbsn}没意思，什么都不想做，不知道该做什么，无聊\end{CJK*}} \\
    \hline
    Fatigue & \makecell[c{p{4cm}}]{fatigue, weariness\\ \begin{CJK*}{UTF8}{gbsn}疲劳，困倦\end{CJK*}} & \makecell[c{p{4cm}}]{tired, exhausted\\ \begin{CJK*}{UTF8}{gbsn}累，没力气\end{CJK*}} \\
    \hline
    Attention & \makecell[c{p{4cm}}]{have difficulty in concentrating\\ \begin{CJK*}{UTF8}{gbsn}难以集中注意力\end{CJK*}} & \\
    \hline
    Self-worth & \makecell[c{p{4cm}}]{self-blame, low self-worth, damaged self-esteem\\ \begin{CJK*}{UTF8}{gbsn}自罪，自我价值感低，自尊心受到打击\end{CJK*}} & \makecell[c{p{4cm}}]{worthless, useless, meaningless, no point\\ \begin{CJK*}{UTF8}{gbsn}一无是处，没用，有什么意义，没有意义\end{CJK*}}\\
    \hline
    Pessimism & \makecell[c{p{4cm}}]{hopeless\\ \begin{CJK*}{UTF8}{gbsn}无望\end{CJK*}} & \\
    \hline
    Sleep Disturbance & \makecell[c{p{4cm}}]{sleep disturbance, excessive sleepiness\\ \begin{CJK*}{UTF8}{gbsn}睡眠困难，嗜睡\end{CJK*}} & \makecell[c{p{4cm}}]{can't sleep, insomnia, tossing and turning\\ \begin{CJK*}{UTF8}{gbsn}睡不好，睡不着，失眠，翻来覆去\end{CJK*}} \\
   \hline
    Weight and Appetite Change & \makecell[c{p{4cm}}]{Increased appetite, decreased appetite, loss of appetite\\ \begin{CJK*}{UTF8}{gbsn}食欲增加，食欲下降，食欲不振\end{CJK*}} & \makecell[c{p{4cm}}]{No appetite, not in the mood to eat, poor appetite\\ \begin{CJK*}{UTF8}{gbsn}没胃口，没什么胃口，胃口不好，饭量\end{CJK*}} \\
    \hline
    Psychomotor retardation &\makecell[c{p{4cm}}]{sluggish thinking\\ \begin{CJK*}{UTF8}{gbsn}思维迟缓\end{CJK*}} & \makecell[c{p{4cm}}]{Mind goes blank\\ \begin{CJK*}{UTF8}{gbsn}脑子一片空白\end{CJK*}} \\
    \hline
    Psychomotor agitation & \makecell[c{p{4cm}}]{Agitation, restlessness, irritability, or excessive talking\\ \begin{CJK*}{UTF8}{gbsn}精神运动性激越，不安，烦躁不安，兴奋或话多\end{CJK*}} & \makecell[c{p{4cm}}]{anxious, mentally unsettled, mind is racing, can't sit still\\ \begin{CJK*}{UTF8}{gbsn}烦躁，静不下心，好像脑子一直在想事情，坐不住\end{CJK*}} \\
    \hline
    Self-harm or Suicide & \makecell[c{p{4cm}}]{suicidal and self-harming thoughts\\ \begin{CJK*}{UTF8}{gbsn}自杀和自伤的想法\end{CJK*}} & \makecell[c{p{4cm}}]{want to die, jump off a building\\ \begin{CJK*}{UTF8}{gbsn}不活，跳楼\end{CJK*}}\\
    \hline
    \end{tabular}
    \caption{The Lexicon of Robot-like Words and Human-like Words}
    \label{tab:robot_human_words}
\end{table*}

%% file: related.tex
\section{Related Works}
\label{sec:related}

\subsection{ChatGPT for Mental Health}
Recently, several studies have assessed the performance of ChatGPT in tasks like depression detection \cite{lamichhane2023evaluation}, emotional conversation \cite{zhao2023chatgpt}, factor detection of mental health conditions, and emotion recognition in conversations \cite{yang2023evaluations}. However, these evaluations were performed on existing datasets using conventional metrics, and did not involve human interaction. What's more, \citet{qin2023read} developed a chat interface using ChatGPT.  
However, they mainly focus on more interpretable and interactive depression detection from social media, while our work focuses on outpatient scenarios, where information should be obtained from conversation, and user experience is the major concern.

\subsection{Doctor Chatbot}
Automatic diagnosis by doctor chatbot has significant practical applications. It enables large-scale screening, alleviates the issue of insufficient medical resources, and provides patients with a more engaged experience than using scales like PHQ-9 \cite{kroenke2001phq}.

While numerous chatbots have been developed to automatically diagnose physical illnesses \cite{Xu2019End, wei2018task}, such chatbots remain relatively uncommon in the mental health domain due to the difficulty in obtaining dialogue data because of ethical concerns. \citet{yao-etal-2022-d4} introduced a depression diagnosis dialogue dataset performed by patient and doctor actors, and a doctor chatbot trained on it. Although the chatbot conduct the diagnostic process correctly, it lacks adequate emotional support and the diagnostic process is inflexible. Another pioneer work \cite{liu2021towards} defines various empathy strategies for mental health support and proposed a meticulously annotated dialogue dataset with these strategies. Recently, \citet{wei2023leveraging} proposed an LLM-based chatbot for information collection, which shares similarities with doctor chatbot, as the latter also need to thoroughly collect the patients' symptoms. 

\subsection{Patient Chatbot}
Recent years, there has been increasing attention to the development of virtual patients for training clinician-patient communication skills \cite{Chaby2022Embodied}. Simulating more lifelike patients can help develop better doctor chatbots \cite{tseng2021transferable}. Additionally, patient chatbots could serve as standardized patients (SPs) in medical education, as currently, actors are hired to portray patients with mental disorders, which is both costly and time-consuming \cite{Gillette2017Cost}. 

Despite this, there are still limited works on developing patient chatbots, and most of them are rule-based\cite{Llanos2021Lessons}. \citet{Duppuy2020Guidelines} provides several guidelines for the design of virtual patient, such as having a reasonable symptomatology and focusing on the abilities needed for psychiatrists (e.g., the virtual patient can show resistance when the doctor ask questions without empathy).

